\documentclass[11pt,a4paper]{article}
\usepackage{authblk}
\usepackage[hyperref]{naaclhlt2018}
\usepackage{times}
\usepackage{latexsym}
\usepackage{graphicx}
\usepackage{url}

\usepackage{url}

\aclfinalcopy 

\title{Learning Patient Representations from Text}

\author[1]{\bf Dmitriy Dligach}
\author[2]{\bf Timothy Miller}
\affil[1]{Loyola University Chicago}
\affil[2]{Boston Children's Hospital and Harvard Medical School}
\affil[1]{\tt ddligach@luc.edu}
\affil[2]{\tt timothy.miller@childrens.harvard.edu}

\date{}

\begin{document}
\maketitle
\begin{abstract}
Mining electronic health records for patients who satisfy a set of predefined criteria is known in medical informatics as phenotyping. Phenotyping has numerous applications such as outcome prediction, clinical trial recruitment, and retrospective studies. Supervised machine learning for phenotyping typically relies on sparse patient representations such as bag-of-words. We consider an alternative that involves learning patient representations. We develop a neural network model for learning patient representations and show that the learned representations are general enough to obtain state-of-the-art performance on a standard comorbidity detection task.
\end{abstract}

\section {Introduction}

Mining electronic health records for patients who satisfy a set of predefined criteria is known in medical informatics as phenotyping. Phenotyping has numerous applications such as outcome prediction, clinical trial recruitment, and retrospective studies. Supervised machine learning is currently the predominant approach to automatic phenotyping and it typically relies on sparse patient representations such as bag-of-words and bag-of-concepts \citep{shivade2013review}. We consider an alternative that involves learning patient representations. Our goal is to develop a conceptually simple method for learning lower dimensional dense patient representations that succinctly capture the information about a patient and are suitable for downstream machine learning tasks. Our method uses cheap supervision in the form of billing codes and thus has representational power of a large dataset. The learned representations can be used to train phenotyping classifiers with much smaller datasets.

Recent trends in machine learning have used neural networks for representation learning, and these ideas have propagated into the clinical informatics literature, using information from electronic health records to learn dense patient representations \citep{choi2016doctor,choi2017gram,Lipton2016,Miotto2016,Nguyen2017,pham2016deepcare}. Most of this work to date has used only codified variables, including ICD (International Classification of Diseases) codes, procedure codes, and medication orders, often reduced to smaller subsets. Recurrent neural networks are commonly used to represent temporality \cite{choi2016doctor,choi2017gram,Lipton2016,pham2016deepcare}, and many methods map from code vocabularies to dense “embedding” input spaces \cite{choi2016doctor,choi2017gram,Nguyen2017,pham2016deepcare}.

One of the few patient representation learning systems to incorporate electronic medical record (EMR) \textit{text} is DeepPatient \cite{Miotto2016}. This system takes as input a variety of features, including coded diagnoses as the above systems, but also uses topic modeling on the text to get topic features, and applies a tool that maps text spans to clinical concepts in standard vocabularies (SNOMED and RxNorm). To learn the representations they use a model consisting of stacked denoising autoencoders. In an autoencoder network, the goal of training is to reconstruct the input using hidden layers that compress the size of the input. The output layer and the input layer therefore have the same size, and the loss function calculates reconstruction error. The hidden layers thus form the patient representation. This method is used to predict novel ICD codes (from a reduced set with 78 elements) occurring in the next 30, 60, 90, and 180 days.

Our work extends these methods by building a neural network system for learning patient representations using text variables only. We train this model to predict billing codes, but solely as a means to learning representations. We show that the representations learned for this task are general enough to obtain state-of-the-art performance on a standard comorbidity detection task. Our work can also be viewed as an instance of transfer learning \cite{pan2010survey}: we store the knowledge gained from a source task (billing code prediction) and apply it to a different but related target task.

\section{Methods}
\subsection{Patient Representation Learning}

The objective of patient representation learning is to map raw text of patient notes to a dense vector that can be subsequently used for various patient-level predictive analytics tasks such as phenotyping, outcome prediction, and cluster analysis. The process of learning patient representations involves two phases: (1) supervised training of a neural network model on a source task that has abundant labeled data linking patients with some outcomes; (2) patient vector derivation for a target task performed by presenting new patient data to the network and harvesting the resulting representations from one of the hidden layers.

In this work, we utilize billing codes as a source of supervision for learning patient vectors in phase 1. Billing codes, such as ICD9 diagnostic codes, ICD9 procedure codes, and CPT codes are derived manually by  medical coders from patient records for the purpose of billing. Billing codes are typically available in abundance in a healthcare institution and present a cheap source of supervision. Our hypothesis is that a patient vector useful for predicting billing codes will capture key characteristics of a patient, making this vector suitable for patient-level analysis.

For learning dense patient vectors, we propose a neural network model that takes as input a set of UMLS concept unique identifiers (CUIs) derived from the text of the notes of a patient and jointly predicts all billing codes associated with the patient. CUIs are extracted from notes by mapping spans of clinically-relevant text (e.g. \emph{shortness of breath}, \emph{appendectomy}, \emph{MRI}) to entries in the UMLS Metathesaurus. CUIs can be easily extracted by existing tools such as Apache cTAKES (http://ctakes.apache.org). Our neural network model (Figure 1) is inspired by Deep Averaging Network (DAN) \cite{iyyer2015deep}, FastText \cite{joulin2016bag}, and continuous bag-of-words (CBOW) \cite{mikolov2013efficient,mikolov2013linguistic} models.

\textbf{Model Architecture}: The model takes as input a set of CUIs. CUIs are mapped to 300-dimensional concept embeddings which are averaged and passed on to a 1000-dimensional hidden layer, creating a vectorial representation of a patient. The final network layer consists of n sigmoid units that are used for joint billing code prediction. The output of each sigmoid unit is converted to a binary (1/0) outcome. The number of units n in the output layer is equal to the number of unique codes being predicted. The model is trained using binary cross-entropy loss function using RMSProp optimizer. Our model is capable of jointly predicting multiple billing codes for a patient, placing it into the family of supervised multi-label classification methods. In our preliminary work, we experimented with CNN and RNN-based architectures but their performance was inferior to the model described here both in terms of accuracy and speed.

\begin{figure}
\begin{center}
\includegraphics[scale=0.22]{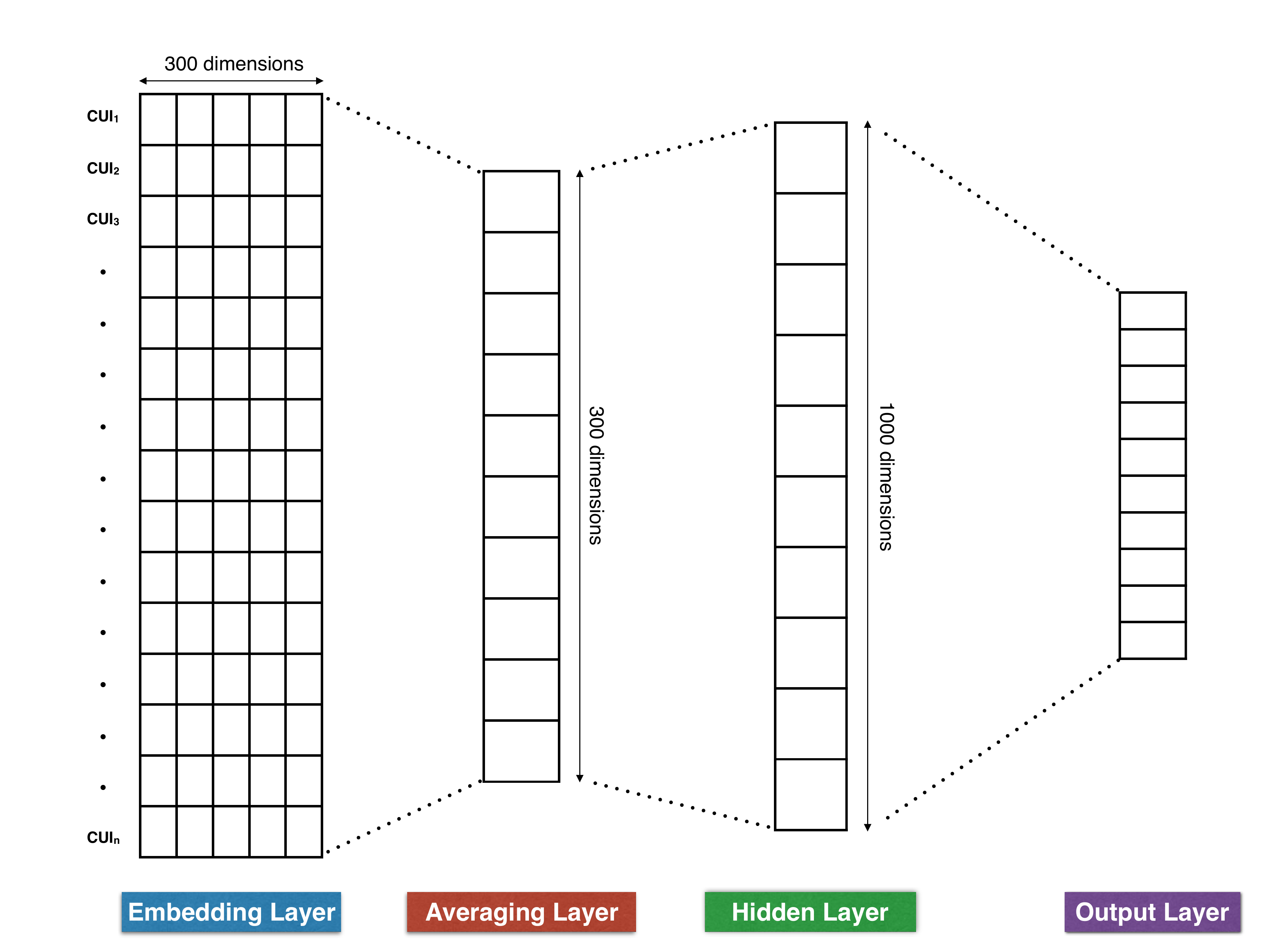}
\caption{Neural network model for learning patient representations from text.}
\end{center}
\end{figure}

Once the model achieves an acceptable level of performance, we can compute a vector representing a new patient by freezing the network weights, pushing CUIs for a new patient through the network, and harvesting the computed values of the nodes in the hidden layer. The resulting 1000-dimensional vectors can be used for a variety of machine learning tasks.

\subsection{Datasets}

For training patient representations, we utilize the MIMIC III corpus \cite{johnson2016mimic}. MIMIC III contains notes for over 40,000 critical care unit patients admitted to Beth Israel Deaconess Medical Center as well as ICD9 diagnostic, procedure, and Current Procedural Terminology (CPT) codes. Since our goal is learning patient-level representations, we concatenate all available notes for each patient into a single document. We also combine all ICD9 and CPT codes for a patient to form the targets for the prediction task. Finally, we process the patient documents with cTAKES to extract UMLS CUIs. cTAKES is an open-source system for processing clinical texts which has an efficient dictionary lookup component for identifying CUIs, making it possible to process a large number of patient documents.

To decrease training time, we reduce the complexity of the prediction task as follows: (1) we collapse all ICD9 and CPT codes to their more general category (e.g. first three digits for ICD9 diagnostic codes), (2) we drop all CUIs that appear fewer than 100 times, (3) we discard patients that have over 10,000 CUIs, (4) we discard all billing codes that have fewer than 1,000 examples. This preprocessing results in a dataset consisting of 44,211 patients mapped to multiple codes (174 categories total). We randomly split the patients into a training set (80\%) and a validation set (20\%) for tuning hyperparameters.

For evaluating our patient representations, we use a publicly available dataset from the Informatics for Integrating Biology to the Bedside (i2b2) Obesity challenge \cite{uzuner2009recognizing}. Obesity challenge data consisted of 1237 discharge summaries from the Partners HealthCare Research Patient Data Repository annotated with respect to obesity and its fifteen most common comorbidities. Each patient was thus labeled for sixteen different categories. We focus on the more challenging \emph{intuitive} task \cite{uzuner2009recognizing,miller2016unsupervised}, containing three label types (\emph{present}, \emph{absent}, \emph{questionable}), where annotators labeled a diagnosis as \emph{present} if its presence could be inferred (i.e., even if not explicitly mentioned). This task involves complicated decision-making and inference.

Importantly, our patient representations are evaluated in sixteen different classification tasks with patient data originating from a healthcare institution different from the one our representations were trained on. This setup is challenging yet it presents a true test of robustness of the learned representations.

\subsection{Experiments}

Our first baseline is an SVM classifier trained with bag-of-CUIs features. Our second baseline involves linear dimensionality reduction performed by running singular value decomposition (SVD) on a patient-CUI matrix derived from the MIMIC corpus, reducing the space by selecting the 1000 largest singular values, and mapping the target instances into the resulting 1000-dimensional space.

Our multi-label billing code classifier is trained to maximize the macro F1 score for billing code prediction on the validation set. We train the model for 75 epochs with a learning rate of 0.001 and batch size of 50. These hyperparameters are obtained by tuning the model's macro F1 on the validation set. Observe that tuning of hyperparameters occurred independently from the target task. Also note that since our goal is not to obtain the best possible performance on a held out set, we are not allocating separate development and test sets. Once we determine the best values of these hyperparameters, we combine the training and validation sets and retrain the model. We train two version of the model: (1) with randomly initialized CUI embeddings, (2) with word2vec-pretrained CUI embeddings.  Pre-trained embeddings are learned using word2vec \cite{mikolov2013efficient} by extracting all CUIs from the text of MIMIC III notes and using the CBOW method with windows size of 5 and embedding dimension of 300.

We then create a 1000-dimensional vector representation for each patient in the i2b2 obesity challenge data by giving the sparse (CUI-based) representation for each patient as input to the ICD code classifier. Rather than reading the classifier's predictions, we harvest the hidden layer outputs, forming a 1000-dimensional dense vector. We then train multi-class SVM classifiers for each disease (using one-vs.-all strategy), building sixteen SVM classifiers. Following the i2b2 obesity challenge, the models are evaluated using macro precision, recall, and F1 scores \cite{uzuner2009recognizing}.

We make the code available
for use by the research community \footnote{\url{https://github.com/dmitriydligach/starsem2018-patient-representations}}.

\section{Results}

\begin{table*}[h!]
\centering
\begin{tabular}{|l||r|r|r||r|r|r||r|r|r|}
\hline
Disease & \multicolumn{3} {c||}{Sparse} & \multicolumn{3} {c||}{SVD} & \multicolumn{3} {c|}{Learned} \\
\cline{2-4} \cline{5-7} \cline{8-10}
& P & R & F1 & P & R & F1 & P & R & F1 \\
\hline
Asthma &	0.894 &	0.736 &	0.787 &	0.888 & 0.854 & 0.870 & 0.910 &	0.920	& 0.915 \\
CAD	& 0.583 &	0.588 &	0.585 &	0.593 & 0.602 & 0.596 & 0.596 &	0.596 &	0.596 \\
CHF &	0.558 &	0.564 &	0.561 &	0.571 & 0.575 & 0.573 & 0.558	& 0.564 &	0.561 \\
Depression &	0.797	& 0.685 &	0.715 & 0.723 & 0.727 & 0.725 &	0.781 &	0.773 &	0.777 \\
Diabetes &	0.859 &	0.853 &	0.856 & 0.611 & 0.624 & 0.617 &	0.907	& 0.919	& 0.913 \\
GERD	& 0.530 &	0.466 &	0.485 &	0.533 & 0.482 & 0.499 & 0.528 &	0.539	& 0.533 \\
Gallstones & 0.814 & 0.640 & 0.678 & 0.747 & 0.721 & 0.732 & 0.645 &	0.663 &	0.653 \\
Gout &	0.975 &	0.811 &	0.871 &	0.955 & 0.834 & 0.882 & 0.928 &	0.910 &	0.919 \\
Hypercholesterolemia & 0.781 & 0.784 & 0.782 & 0.789 & 0.793 & 0.790 & 0.865 &	0.868 &	0.866 \\
Hypertension &	0.680 &	0.650 &	0.662 & 0.711 & 0.763 & 0.728 &	0.825 &	0.879 &	0.847 \\
Hypertriglyceridemia &	0.933 &	0.679 &	0.748 & 0.580 & 0.610 & 0.591 &	0.604 &	0.650 &	0.621 \\
OA &	0.514 &	0.448 &	0.466 & 0.479 & 0.442 & 0.454 &	0.511	& 0.508	& 0.510 \\
OSA &	0.596 &	0.511 &	0.542 & 0.626 & 0.568 & 0.592 &	0.611 &	0.618 &	0.615 \\
Obesity &	0.825 &	0.791 &	0.798 & 0.883 & 0.844 & 0.853 &	0.872 &	0.873 &	0.872 \\
PVD &	0.594 &	0.542 &	0.564 &	0.599 & 0.557 & 0.576 & 0.568 &	0.599 &	0.582 \\
Venous Insufficiency &	0.797	& 0.649 &	0.694 & 0.669 & 0.757 & 0.700 &	0.638 &	0.717 &	0.665 \\
\hline
Average &	0.733 &	0.650 &	0.675 &	0.685 & 0.672 & 0.674 & 0.709	& 0.725 &	0.715 \\
\hline
\end{tabular}
\caption{\label{table:results} Comorbidity challenge results (intuitive task). SVM trained using sparse representations (bag-of-CUIs) is compared to SVM trained using SVD-based representations and learned dense patient representations.}
\end{table*}

Our billing code classifier achieves the macro F1 score on the source task (billing code prediction) of 0.447 when using randomly initialized CUI embeddings and macro F1 of 0.473 when using pre-trained CUI embeddings. This is not directly comparable to existing work because it is a unique setup; but we note that this is likely a difficult task because of the large output space. However, it is interesting to note that pre-training CUI embedding has a positive relative impact on performance.

Classifier performance for the target phenotyping task is shown in Table \ref{table:results}, which shows the performance of the baseline SVM classifier trained using the standard bag-of-CUIs approach (Sparse), the baseline using 1000-dimensional vectors obtained via dimensionality reduction (SVD), and our system using dense patient vectors derived from the source task. Since a separate SVM classifier was trained for each disease, we present classifier performance for each SVM model.

Both of our baseline approaches showed approximately the same performance (F1=0.675) as the best reported i2b2 system \cite{solt2009semantic} (although they used a rule-based approach). Our dense patient representations outperformed both baseline approaches by four percentage points on average (F1=0.715). The difference is statistically significant (t-test, p=0.03).

Out of the sixteen diseases, our dense representations performed worse (with one tie) than the sparse baseline only for three: gallstones, hypertriglyceridemia, venous insufficiency. The likely cause is the scarcity of positive training examples; two of these diseases have the smallest number of positive training examples.

\section{Discussion and Conclusion}

For most diseases and on average our dense patient representations outperformed sparse patient representations. Importantly, patient representations were learned from a task (billing code prediction) that is different from the evaluation task (comorbidity prediction), presenting evidence that useful representations can be derived in this transfer learning scenario.

Furthermore, the data from which the representations were learned (BI medical center) and the evaluation data (Partners HealthCare) originated from two different healthcare institutions providing evidence of robustness of our patient representations.

Our future work will include exploring the use of other sources of supervision for learning patient representations, alternative neural network architectures, tuning the learned patient representations to the target task, and evaluating the patient representations on other phenotyping tasks.

\section*{Acknowledgments}

The Titan X GPU used for this research was donated by the NVIDIA Corporation.
Timothy Miller's effort was supported by National Institute of General Medical Sciences (NIGMS) of the National Institutes of Health under award number R01GM114355.
The content is solely the responsibility of the authors and does not necessarily represent the official views of the National Institutes of Health.

\bibliographystyle{acl_natbib}
\bibliography{references}

\end{document}